\pdfoutput=1
\documentclass[letterpaper]{article} 
\usepackage{aaai24}  
\usepackage{times}  
\usepackage{helvet}  
\usepackage{courier}  
\usepackage[hyphens]{url}  
\usepackage{graphicx} 
\usepackage{natbib}  
\usepackage{caption} 
\usepackage{algorithm}
\usepackage{algorithmic}
\usepackage{tabularx} 
\usepackage{amsthm} 
\usepackage{multirow} 
\usepackage{graphics}
\usepackage{booktabs} 
\usepackage{color} 
\usepackage{amsmath}
\usepackage{amssymb}

\usepackage{newfloat}
\usepackage{listings}
\DeclareCaptionStyle{ruled}{labelfont=normalfont,labelsep=colon,strut=off} 
\floatstyle{ruled}
\newfloat{listing}{tb}{lst}{}
\floatname{listing}{Listing}
\title{Tuning-Free Inversion-Enhanced Control for Consistent Image Editing}
\author{
    Xiaoyue Duan\textsuperscript{\rm 1,2}\equalcontrib\thanks{Work done during an internship in Meituan.}
    Shuhao Cui\textsuperscript{\rm 2}\equalcontrib,
    Guoliang Kang\textsuperscript{\rm 1,4},
    Baochang Zhang\textsuperscript{\rm 1,3,4,5},\\
    Zhengcong Fei\textsuperscript{\rm 2},
    Mingyuan Fan\textsuperscript{\rm 2},
    Junshi Huang\textsuperscript{\rm 2}\thanks{Corresponding author.}
}
\affiliations{
    \textsuperscript{\rm 1}School of Automation Science and Electrical Engineering, Beihang University, China
    \textsuperscript{\rm 2}Meituan\\
    \textsuperscript{\rm 3}Hangzhou Research Institute, Beihang University, China
    \textsuperscript{\rm 4}Zhongguancun Laboratory, Beijing, China\\
    \textsuperscript{\rm 5}Nanchang Institute of Technology, Nanchang, China

    \{LittleMoon, bczhang\}@buaa.edu.cn, \{hassassin1621, kgl.prml, feizhengcong, fanmingyuan01, junshi.huang\}@gmail.com
}

\usepackage{bibentry}

\begin{document}

\maketitle

\begin{abstract}
Consistent editing of real images is a challenging task, as it requires performing non-rigid edits (\emph{e.g.}, changing postures) to the main objects in the input image without changing their identity or attributes. To guarantee consistent attributes, some existing methods fine-tune the entire model or the textual embedding for structural consistency, but they are time-consuming and fail to perform non-rigid edits. Other works are tuning-free, but their performances are weakened by the quality of Denoising Diffusion Implicit Model (DDIM) reconstruction, which often fails in real-world scenarios. In this paper, we present a novel approach called Tuning-free Inversion-enhanced Control (TIC), which directly correlates features from the inversion process with those from the sampling process to mitigate the inconsistency in DDIM reconstruction. Specifically, our method effectively obtains inversion features from the key and value features in the self-attention layers, and enhances the sampling process by these inversion features, thus achieving accurate reconstruction and content-consistent editing. To extend the applicability of our method to general editing scenarios, we also propose a mask-guided attention concatenation strategy that combines contents from both the inversion and the naive DDIM editing processes. Experiments show that the proposed method outperforms previous works in reconstruction and consistent editing, and produces impressive results in various settings.
\end{abstract}

\section{Introduction}

\begin{figure}[t]
    \centering
    \includegraphics[width=\linewidth]{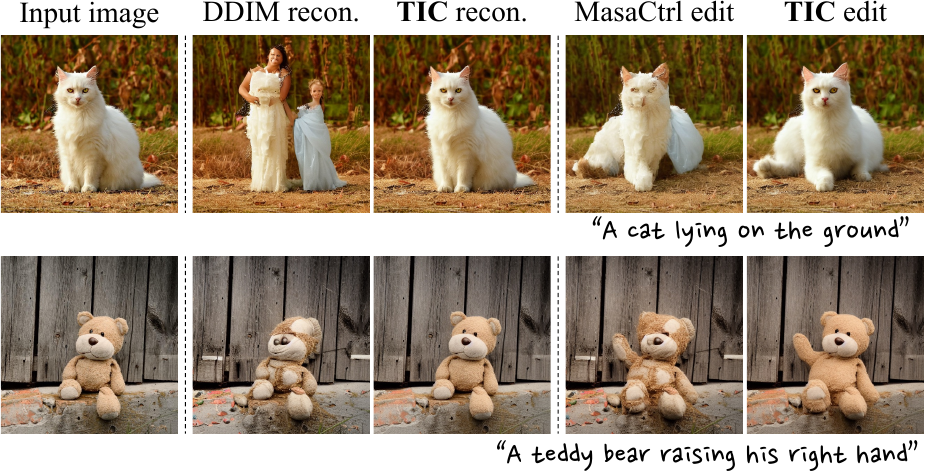}
    \vspace{-6.5mm}
    \caption{Naive DDIM reconstruction may result in bad cases (col. 2). Therefore, querying the contents from the naive DDIM reconstruction process during editing (\emph{i.e.}, MasaCtrl) lead to unreasonable editing results (col. 4). By enhancing the self-attention layer with contents from inversion, our method accurately reconstruct the input image (col. 3), and thus achieve content-consistent editing (col. 5).}
    \label{fig: reconstruction_and_editing_comoparison_with_ddim}
    \vspace{-4.5mm}
\end{figure}

In recent years, remarkable progress in text-to-image (T2I) generation has been witnessed, as evidenced by the impressive results by powerful models~\cite{ramesh2021zero, ding2022cogview2, ramesh2022hierarchical}. These large-scale T2I models (\emph{e.g.}, Stable Diffusion~\cite{rombach2022high}) are capable of generating diverse and high-quality images that match the given text descriptions. Moreover, by leveraging T2I models, we can also perform text-guided image editing, as demonstrated by recent works~\cite{nichol2021glide, parmar2023zero}. These works involve various forms of text guidance and task settings, which highlight the need to address the challenges and opportunities in  field.

In the realm of text-guided image editing, a crucial and  practical task is to achieve consistent image editing. Consistent image editing, as demonstrated in recent works such as~\cite{meng2021sdedit,hertz2022prompt}, involves preserving the identity of objects and background details in the input image while modifying only certain non-rigid attributes of the objects (\emph{e.g.}, changing posture). Nevertheless, this requirement remains challenging, as existing text-guided editing methods can hardly solve the problems of consistent editing effectively. For example, some methods~\cite{brooks2022instructpix2pix, tumanyan2022plug} are capable of preserving the structure or layout of the input image when performing style transfer or object replacement, but they can barely preserve the content or identity of the object, resulting in inconsistent editing. Imagic~\cite{kawar2022imagic} can preserve the original attributes of the object, which allows for non-rigid editing. Yet, it requires fine-tuning the entire T2I model and optimizing the textual embedding for each input image, which is not acceptable for real-world applications in terms of efficiency. 

To achieve efficient consistent image editing, subsequent
works~\cite{mokady2022null,dong2023prompt,han2023improving} avoid fine-tuning the entire model, and learn the structural information of the input image by tuning either the unconditional or the conditional embedding of the classifier-free guidance based on a pivotal latent trajectory obtained through DDIM inversion~\cite{ho2020denoising}. However, tuning the embedding for each edit is still time-consuming, and cannot achieve complex non-rigid editing. Recent works, \emph{e.g.}, MasaCtrl~\cite{cao2023masactrl},
achieve consistent editing without fine-tuning any part of the model, but it may introduce artifacts in real-image editing scenarios, and its performance is largely constrained by DDIM reconstruction quality, as shown in Fig.~\ref{fig: reconstruction_and_editing_comoparison_with_ddim}.

In this paper, we present a novel approach called Tuning-free Inversion-enhanced Control (TIC) for consistent image editing. Our approach is based on the theoretical analysis of the reconstruction error between the DDIM inversion and sampling processes. We find that the reconstruction error is mainly caused by the bias in the predicted noises resulted from an inaccurate timestep approximation assumption.
To address this issue, we propose to enhance the DDIM sampling process by incorporating features from the DDIM inversion process. Specifically, we enhance the self-attention layers by replacing the key and value features in the sampling process with the corresponding features in the inversion process, thus allowing the model to focus on important pixels in the image features.
Our method is tuning-free, and outperforms existing editing methods in both image reconstruction and consistent editing of real images, as demonstrated in Fig.~\ref{fig: reconstruction_and_editing_comoparison_with_ddim}.

To extend our method to more general and diverse editing scenarios, we further propose a mask-guided attention concatenation strategy, which achieves a good balance between fidelity and editability by querying contents from both the inversion process and the naive DDIM editing process in the mask-guided editing areas. We also demonstrate the effectiveness of our method by integrating it into controllable diffusion models to further enhance the structural layout of the input image. Experiments show the effectiveness and applicability of our method in various settings.

Overall, our contributions can be summarized as: 
\begin{itemize}
    \item We conduct theoretical analysis on the reconstruction error of DDIM. Based on the analysis, we propose Tuning-free Inversion-enhanced Control (TIC) to achieve accurate reconstruction and consistent editing of real images.
    \item We extend TIC to more general editing settings by proposing a mask-guided attention concatenation strategy, which achieves a good balance between fidelity and editability. We also demonstrate its effectiveness when integrated with controllable diffusion models.
    \item We demonstrate the versatility and applicability of the proposed method by conducting experiments under both qualitative and quantitative settings.
\end{itemize}

\section{Related Work}

\textbf{Text-to-image generation.} Generating images by text descriptions are mainly based on architectures of Generative Adversarial Networks (GANs)~\cite{reed2016generative, zhang2018stackgan++, brock2018large, tao2022df}, auto-regressive generation~\cite{ding2021cogview} and diffusion models~\cite{song2019generative, nichol2021improved,ramesh2022hierarchical}.
Early methods based on GANs~\cite{xu2018attngan, zhang2021cross,zhou2022tigan} align the text descriptions and image contents through multi-modal vision-language learning, but can only achieve impressive results on specific domains. By adopting large-scale models and datasets, auto-regressive generation~\cite{ramesh2021zero,ding2022cogview2,yu2022parti} obtain powerful results for open-domain text descriptions. More recently, diffusion models~\cite{song2020denoising,ho2020denoising, dhariwal2021diffusion,gu2022vector,ho2022classifier} achieve state-of-the-art synthesis results in terms of image quality and diversity. Conditioned on the text prompt, various text-to-image diffusion models~\cite{nichol2021glide,ramesh2022hierarchical,zeng2023ipdreamer,zeng2023controllable} can synthesize image contents highly consistent with the textual description. 

\textbf{Text-guided image editing.} Text-guided image editing is a challenging task that involves manipulating images based on natural language descriptions. Previous methods based on GANs~\cite{nam2018text, li2020manigan, xia2021tedigan, patashnik2021styleclip,pan2023draggan}, or auto-regressive models~\cite{crowson2022vqgan} have achieved some success on domain-specific datasets, but their applicability are limited. Recently, the development of diffusion models provide a more flexible space and a more efficient way for editing, while following a simpler setup~\cite{meng2021sdedit}. Some works~\cite{nichol2021glide, avrahami2022blended} leverage extra masks to edit specific regions of the image, while others~\cite{kim2022diffusionclip, brooks2022instructpix2pix} can edit global aspects of the image by directly modifying the text prompt.

\textbf{Consistent image editing. }Consistent image editing refers to the process of editing images without altering their main components. Imagic~\cite{kawar2022imagic} allows for various non-rigid editing by directly modifying the prompts. \citet{hertz2022prompt} utilizes cross-attention or spatial features to edit both global and local aspects of the image by modifying the text prompt. Later, \citet{mokady2022null,dong2023prompt} use an initial DDIM inversion as an anchor for optimization, which only tunes the prompt embeddings used in classifier-free guidance. Recently, MasaCtrl~\cite{cao2023masactrl} combines the contents from the source image and the layout synthesized from text prompt to synthesize or edit the desired image. However, these methods either require fine-tuning, which is time-consuming and fail to perform non-rigid edits, or are highly constrained by the DDIM reconstruction quality. For fast and effective editing, a tuning-free method with high reconstruction quality is in need.

\begin{figure*}[t]
    \centering
    \includegraphics[width=0.88\textwidth]{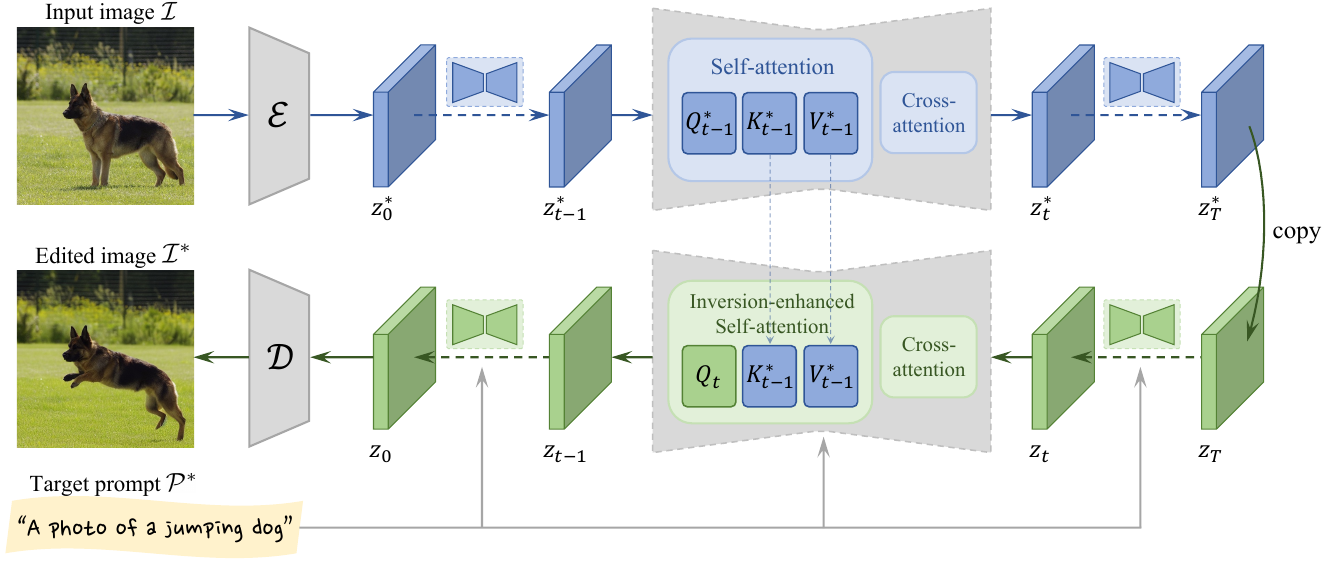}
    \vspace{-3mm}
    \caption{Overview of Tuning-free Inversion-enhanced Control (TIC). Our method first performs DDIM inversion on a given real image to obtain a series of features in the self-attention layers (1st row). These features from inversion contain valuable texture information of the input, which we adopt to enhance the self-attention layers to perfectly reconstruct the input, and achieve non-rigid and content-consistent image editing with the guidance of the target prompt (2nd row).}
    \label{fig: main_fig}
        \vspace{-3.5mm}
\end{figure*}

\section{Methodology}

Given a real image $\mathcal{I}$ and a target prompt $\mathcal{P}^*$, consistent image editing aims to perform non-rigid edits to $\mathcal{I}$ (\emph{e.g.}, changing postures) to make its visual content comply with the textual content in $\mathcal{P}^*$, while preserving the original texture and identity of the main components in $\mathcal{I}$. To preserve the texture and identity, we believe that an accurate reconstruction of the input is a basic guarantee for consistent image editing.



In this section, we first analyze the limitations of DDIM reconstruction in Latent Diffusion Models (LDMs), which may result in failed reconstruction of the input. Based on the analysis, we then propose Tuning-free Inversion-enhanced Control (TIC) (see Fig.~\ref{fig: main_fig} for an overview), which adopts features from inversion to enhance the sampling process. With TIC, we can perfectly reconstruct the input image, and thus perform non-rigid consistent image editing. Finally, we introduce the extensions of TIC for more typical editing, including a mask-aware attention concatenation strategy, and an integration to controllable diffusion models.

\subsection{Limitations of DDIM Reconstruction}
\label{sec: 3.1}

Following the framework of Latent Diffusion Models~\cite{rombach2022high},
a pre-trained encoder $\mathcal{E}$ maps the input image $\mathcal{I}$ to a latent representation $z_0$, and the input image can be then reconstructed using a decoder $\mathcal{D}$, \emph{i.e.}, $\mathcal{I}\!\approx\!\mathcal{D}(\mathcal{E}(\mathcal{I}))$, which can be regarded as the upper bound for image reconstruction. 

The deterministic DDIM sampling~\cite{song2020denoising} adopts the following denoising process in LDMs:
\begin{equation}
    \label{E4}
    \frac{z_{t-1}}{\sqrt{\alpha_{t-1}}}=\frac{z_t}{{\sqrt{\alpha_{t}}}}+ \left(\sqrt{\frac{1}{\alpha_{t-1}}-1} - \sqrt{\frac{1}{\alpha_{t}}-1}\right)\cdot\epsilon_{t},
\end{equation}
where $\epsilon_t$ denotes the noise prediction process at timestep $t$. A random Gaussian noise $z_T$ is gradually removed to generate an image latent $z_0$ by applying Eqn.~\ref{E4} for $T$ steps. However, as the Gaussian noise is randomly sampled, the generated image by DDIM sampling can be far different from the input one. For better image reconstruction, DDIM inversion is suggested to reverse DDIM sampling, based on the assumption that the ordinary differential equation (ODE) process can be reversed within the limit of small steps:
\begin{equation}
    \label{E6}
    \frac{z_t^*}{{\sqrt{\alpha_{t}}}}= \frac{z_{t-1}^*}{{\sqrt{\alpha_{t-1}}}}
    + \left(\sqrt{\frac{1}{\alpha_{t}}-1} - \sqrt{\frac{1}{\alpha_{t-1}}-1}\right)\cdot\epsilon_{{t-1}}^*,
\end{equation}
where the superscript $^*$ denotes features in the inversion process.
However, in practice, a slight error is incorporated at each timestep $t$.

Naive DDIM reconstruction first performs DDIM inversion in Eqn.~\ref{E6} for $T$ steps to obtain a pivotal trajectory of latents $\{z_t^*\}_{t=0}^T$. Then, starting from $z_T\!=\!z_T^*$, DDIM sampling is performed based on Eqn.~\ref{E4} to obtain the reconstructed trajectory $\{z_t\}_{t=T}^0$.
This naive DDIM inversion and reconstruction process is known to provide a rough approximation of the original image~\cite{song2020denoising} according to the assumption of the ODE process mentioned above. However, with a larger value of $T$ (\emph{e.g.}, $T\!=\!50$), the errors may be accumulated, leading to failed reconstructions. 

To analyze the error in each reconstruction step (\emph{i.e.}, from $t$ to $t\!-\!1$), we assume the starting latent is the same, \emph{i.e.}, $z_t\!=\!z_t^*$. Then, based on Eqn.~\ref{E4} and~\ref{E6}, the reconstruction error between $z_{t-1}$ and $z_{t-1}^*$ can be calculated as:
\begin{equation}
    \label{eq: reconstructed error at each step}
    \frac{z_{t-1}\!-\!z_{t-1}^*}{\sqrt{\alpha_{t-1}}}=\left(\!\sqrt{\frac{1}{\alpha_{t-1}}-1}-\sqrt{\frac{1}{\alpha_t}-1}\right)\cdot\left(\epsilon_t-\epsilon_{t-1}^*\right),
\end{equation}
where $\alpha_t$ is a fixed value with certain $t$. Then the reconstruction error at step $t$ can be formulated as:
\begin{equation}
    \label{eq: C reconstructed error at each step}
    z_{t-1}-z_{t-1}^*=\mathcal{C}_t\cdot\|\epsilon_t-\epsilon_{t-1}^*\|,
\end{equation}
where $\mathcal{C}_t$ is a constant at step $t$. This indicates that the main source of reconstruction error comes from the difference between the predicted noise in the inversion process $\epsilon_{t-1}^*$ and that in the sampling process $\epsilon_t$ at each step. This difference is resulted from 
an inaccurate timestep approximation assumption from $t$ to $t-1$. Therefore, reducing the error between the predicted noises at each step is the key to accurate reconstruction. 

\subsection{Tuning-Free Inversion-Enhanced Control}
\label{sec: 3.2}

Since the noise prediction is conducted by the U-Net $\epsilon_\theta$, during the inversion process, $\epsilon_{t-1}^*$ can be calculated as $\epsilon_{t-1}^* = \epsilon_\theta(z_{t-1}^*, t-1, \phi)$, where $z_{t-1}^*$, $t-1$ and $\phi$ denote the latent input, the timestep and a null-text input, respectively. 
To reduce the error between $\epsilon_t$ and $\epsilon_{t-1}^*$ at each step, we aim to obtain a sampling trajectory $\{z_t\}_{t=T}^0$ close to the pivotal trajectory $\{z_t^*\}_{t=0}^T$ for accurate reconstruction. To this end, we propose Tuning-free Inversion-enhanced Control (TIC), which utilizes the features from the inversion process to enhance the sampling process and achieve a precise reconstruction of the original image as: 
\begin{equation}
\epsilon_t = \epsilon_\theta(z_t,t,\mathcal{C};z_{t-1}^*),
\end{equation}
where we introduce an extra feature from the inversion process $z_{t-1}^*$ into the sampling process to reduce the difference between $\epsilon_t$ and $\epsilon_{t-1}^*$, as
similar values of $\epsilon_t$ and $\epsilon_{t-1}^*$ ensure a more accurate reconstruction of the input.

To determine which specific feature to adopt for TIC,
we analyze the internal structure of the U-Net $\epsilon_\theta$. U-Net is composed of stacked convolutional residual blocks~\cite{he2016deep} and transformer blocks~\cite{vaswani2017attention}. More specifically, each transformer block mainly consists of a self-attention layer and a cross-attention layer. The attention mechanism in U-Net is known to play a significant role in capturing long-term dependencies and contextual information in the input sequence. The self-attention layers, particularly, allow the model to focus on important pixels of the image features, and thus can provide indications on how the noise should be \textit{added} (in the inversion process) or \textit{removed} (in the sampling process).
Therefore, to introduce an extra input feature $z_{t-1}^*$ into $\epsilon_\theta$, we propose to adopt features from the self-attention layers.

\begin{algorithm}[t]
%

    \caption{Tuning-free Inversion-enhanced Control (TIC) for Consistent Image Editing}
    
    \label{alg: TIC}
        \textbf{Input:} A real image $\mathcal{I}$, a target prompt $\mathcal{P}^*$, and the start timestep index $t_0$ and layer index $l_0$ of TIC\\
        \textbf{Output:} The
        edited image $\mathcal{I}^*$
        \begin{enumerate}
            \item $KV_{\text{list}}=[], z_0^*=\mathcal{E}(\mathcal{I})$
            \item \textbf{for} $t = 0, 1, ..., T-1$ \textbf{do}:
                \item ~~~~$\epsilon^*, \{Q_{t}^*, K_{t}^*, V_{t}^*\}\leftarrow\epsilon_\theta(z_t^*,t,\phi)$
                \item ~~~~$KV_{\text{list}}[t]\leftarrow\{K_{t}^*, V_{t}^*\}$
                \item ~~~~$z_{t+1}^* \leftarrow  \text{Inverse}(z_t^*, \epsilon^*)$
            \item \textbf{end for}
            \item $z_T=z_T^*$
            \item \textbf{for} $t = T, T-1, ..., 1$ \textbf{do}:
            \item ~~~~$\{K_{t-1}^*,V_{t-1}^*\}\leftarrow KV_{\text{list}}[t-1]$
            \item ~~~~$\{Q_t,K_t,V_t\}\leftarrow\epsilon_\theta(z_t,t,\mathcal{P}^*)$
            \item ~~~~\textbf{if} $t>t_0$ \textbf{and} $l>l_0$:
            \item ~~~~~~~~$\epsilon_t=\epsilon_\theta(z_t,t,\mathcal{P}^*;\{Q_t,K_{t-1}^*,V_{t-1}^*\})$
            \item ~~~~\textbf{else}:
            \item ~~~~~~~~$\epsilon_t=\epsilon_\theta(z_t,t,\mathcal{P}^*;\{Q_t,K_t,V_t\})$
            \item ~~~~$z_{t-1}\leftarrow\text{Sample}(z_t,\epsilon_t)$
            \item \textbf{end for}
            \item \textbf{return} $\mathcal{I}^*=\mathcal{D}(z_0)$
        \end{enumerate}

\end{algorithm}

In self-attention layers, the query, key and value features (denoted as $Q$, $K$ and $V$, respectively) are projected from the spatial features. For convenience, we denote the corresponding self-attention features at timestep $t$ as $(Q_{t}^*,K_{t}^*,V_{t}^*)$ for the inversion process, and $(Q_{t},K_{t},V_{t})$ for the sampling process. 
Inspired by the \textit{cross-attention layers}, where the key and value features receive extra information from the text inputs to achieve feature extraction and fusion, we believe that the key and value features in the \textit{self-attention layers} can also receive additional texture information from the input feature in the inversion process, \emph{i.e.}, $z_{t-1}^*$. Since $(K_{t-1}^*,V_{t-1}^*)$ are self-attention features obtained with $z_{t-1}^*$ as the input, they can be directly adopted to enhance the self-attention layers during sampling. Thus, we
directly replace the key and value features $(K_{t},V_{t})$ in the sampling process with the corresponding features $(K_{t-1}^*,V_{t-1}^*)$ in the inversion process, resulting in inversion-enhanced self-attention layers.
Accordingly, the noise can be then calculated as:
\begin{equation}
\label{eq: final mechanism}
\epsilon_t=\epsilon_\theta(z_t,t,\mathcal{C};\{Q_t,K_{t-1}^*,V_{t-1}^*\}),
\end{equation}
where $\mathcal{C}$ is the null-text input $\phi$ for reconstruction, and the target prompt $\mathcal{P}^*$ for consistent editing.
With this feature replacement, the whole framework of TIC is implemented as shown in Fig.~\ref{fig: main_fig}.

We summarize our algorithm for consistent editing in Alg.~\ref{alg: TIC}. We first perform DDIM inversion based on Eqn.~\ref{E6} for $T$ steps to obtain the pivotal trajectory $\{z_t^*\}_{t=0}^T$. During the process, we save the key and value features $(K_t^*,V_t^*)$ in the self-attention layers at each step. Then, starting with $z_T\!=\!z_T^*$, the sampling process is performed with the guidance of $\mathcal{P}^*$ based on Eqn.~\ref{E4} to generate the edited image. During the sampling process, the key and value features $(K_t,V_t)$ in the self-attention layers are replaced with the corresponding features $(K_{t-1}^*,V_{t-1}^*)$ from the inversion process for each step. Note that we do NOT perform inversion-enhanced self-attention for all layers or all denoising steps, since such operation in the early steps or the shallow layers of the U-Net disrupts the layout formation of the target image. Following~\cite{cao2023masactrl}, the proposed inversion-enhanced self-attention is only performed when $l\!>\!l_0$ and $t\!>\!t_0$,
where $l_0$ and $t_0$ are the start timestep index and layer index for TIC, respectively.


The proposed TIC can achieve accurate reconstruction of the input image without the need for any fine-tuning, effectively solving the problems of naive DDIM reconstruction. Based on this, TIC successfully performs non-rigid edits with the guidance of the given texts, while maintaining high content consistency with the input.

\begin{table*}[t]
\centering
\small
\setlength{\tabcolsep}{1.72mm}{
\begin{tabular}{cccccccc}
\toprule[1pt]
     & \begin{tabular}[c]{@{}c@{}}VAE recon.\\ (upper bound)\end{tabular} & DDIM recon.   & \begin{tabular}[c]{@{}c@{}}NTI recon.\\ (iter=250)\end{tabular} & \begin{tabular}[c]{@{}c@{}}NTI recon.\\ (iter=500)\end{tabular} & \begin{tabular}[c]{@{}c@{}}PTI recon.\\ (iter=250)\end{tabular} & \begin{tabular}[c]{@{}c@{}}PTI recon.\\ (iter=500)\end{tabular} & \textbf{\begin{tabular}[c]{@{}c@{}}TIC recon.\\ (ours)\end{tabular}} \\ \midrule[0.5pt]
PSNR ($\uparrow$) & \textbf{27.17}                                                     & 22.95         & 25.70                                                           & 26.69                                                            & 26.71                                                           & 26.92                                                            & \textbf{27.11}                                                        \\
SSIM ($\uparrow$) & \textbf{0.7886}                                                    & 0.6840        & 0.7631                                                          & 0.7797                                                           & 0.7810                                                          & 0.7844                                                           & \textbf{0.7864}                                                       \\
Time ($\downarrow$) & -                                                                  & \textbf{5.56} & 97.86                                                           & 149.34                                                           & 89.89                                                           & 134.53                                                           & \textbf{5.56}                                                         \\ \bottomrule[1pt]
\end{tabular}}
\vspace{-3mm}
\caption{Reconstruction quality (measured by PSNR and SSIM) and time (\textit{s} per image) of different methods. The proposed TIC outperforms all other methods in reconstruction quality, and is much more efficient than the tuning-based methods (\emph{i.e.}, Null-Text Inversion (NTI) and Prompt-Tuning Inversion (PTI)) under different number of iterations.
}
\vspace{-3mm}
\label{table: reconstruction PSNR}
\end{table*}

\subsection{Extensions of TIC}
\label{sec: 4.4}

\textbf{Mask-guided TIC with attention concatenation.} The proposed TIC ensures fidelity to the input, but as it only queries contents from the inversion of the input, it can hardly generate new contents that do not exist in $\mathcal{I}$ (\emph{e.g.}, turn the dog in $\mathcal{I}$ into a cat). This is not friendly for general editing, which often requires new contents (\emph{e.g.}, object replacement). On the contrary, naive DDIM queries almost no content from inversion, and it can generate new contents beyond the original image with the guidance of the prompt. Therefore, we propose to concatenate the key and value features in DDIM sampling process with the corresponding features in the inversion process, \emph{i.e.}, $[K_{t};K_{t-1}^*]$ and $[V_{t};V_{t-1}^*]$, with $[\cdot;\cdot]$ as the concatenation operation. Then the self-attention is formulated as $(Q_{t},[K_{t};K_{t-1}^*],[V_{t};V_{t-1}^*])$. In this way, the model queries both the content faithful to the target prompt, and the content reconstructing the details of the input.

In practice, we aim to edit the desired parts of the image, while preserving the details of other parts. Inspired by previous works~\cite{hertz2022prompt}, we adopt the cross-attention maps to create a binary mask that distinguishes the parts to be edited from the parts to be preserved. Specifically, at each step $t$, we average the cross-attention maps with the spatial resolution of $16\!\times\!16$ across all heads and layers, resulting in a map $A_t\!\in\!\mathbb{R}^{16\times16\times N}$, where $N$ is the number of textual tokens of $\mathcal{P}^*$. We then obtain an averaged cross-attention map for the tokens related to the objects or parts that we want to edit, which is then binarized to obtain the mask. For the editing parts, we adopt the attention concatenation strategy proposed above to query contents from both inversion and naive DDIM editing, balancing fidelity and editability. For other parts that do not require editing, we only query contents from inversion to preserve original details of the input.

\textbf{Integration to controllable diffusion models. }Our method can also be easily integrated with existing controllable image synthesis methods (\emph{e.g.}, ControlNet~\cite{zhang2023adding} and T2I-Adapter~\cite{mou2023t2i}) to better preserve the layout and structure of the input for general editing. Specifically, we first obtain the controllable image map (\emph{e.g.}, depth map) of the input. Then, following the same pipeline as ControlNet, we integrate the controllable image features into the editing process of the mask-guided TIC mentioned above. Through this approach, we achieve more general editing while almost completely preserving the layout and shape of the input. We demonstrate the effectiveness of this combination in the following experiment part.

\begin{figure*}[t]
    \centering
    \includegraphics[width=0.95\textwidth]{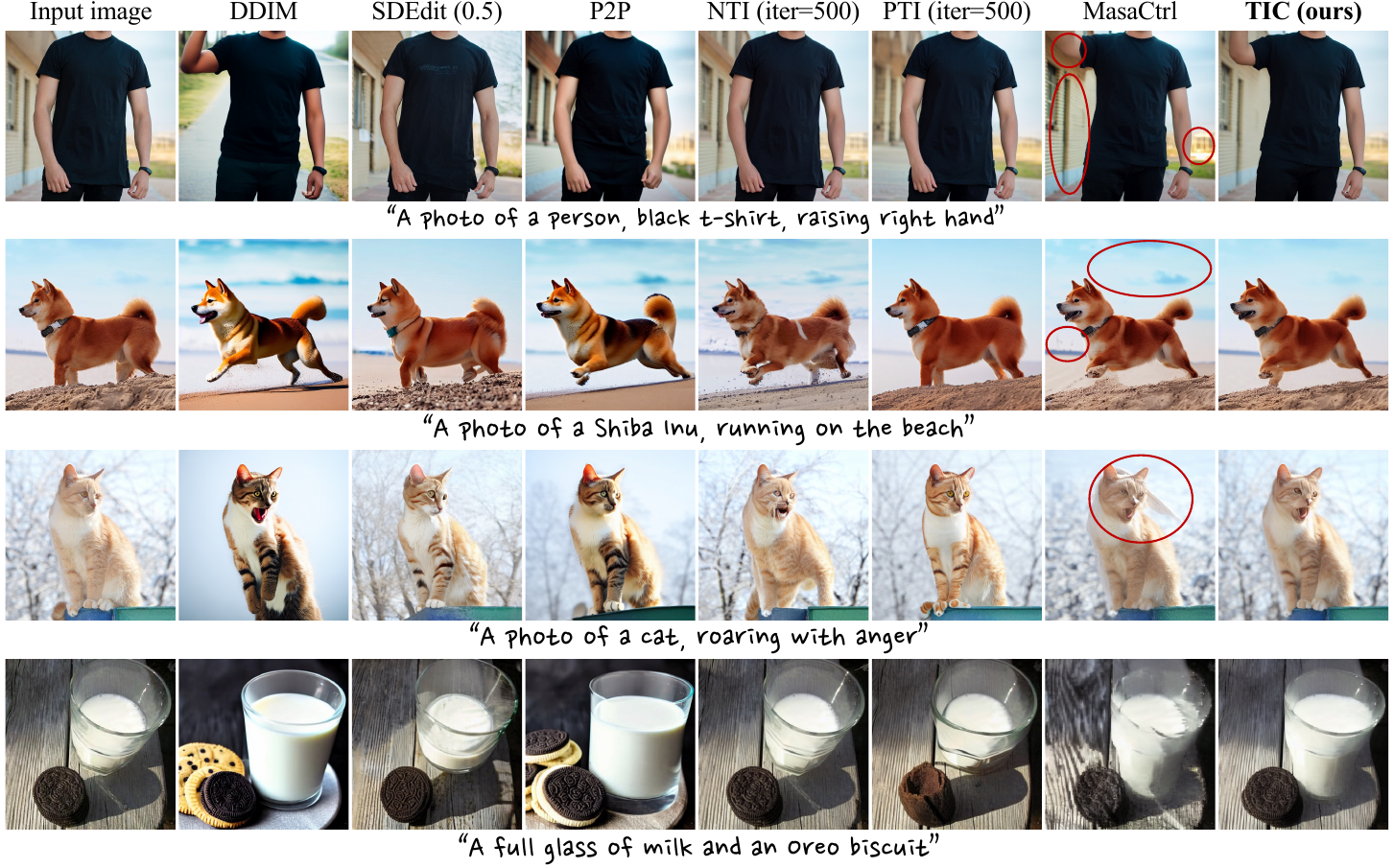}
    \vspace{-2.5mm}
    \caption{Consistent image editing results of different editing methods on real images. Compared to other methods, our method can perform non-rigid edits without introducing artifacts into the image, while maintaining content consistency.}
    \label{fig: consistent_all}
        \vspace{-3mm}
\end{figure*}

\section{Experiments}
\label{sec: experiments}



\subsection{Experiment Setup}
\label{sec: 4.1 setup}

\textbf{Implementation details.} We adopt the text-conditional Latent Diffusion Model~\cite{rombach2022high} (also known as Stable Diffusion) with the publicly available checkpoint v1.4. For the DDIM schedule, we perform both inversion and sampling for 50 steps, and retain the original hyper-parameter choices of Stable Diffusion. The classifier-free guidance (CFG) scale is set to 7.5 for editing. The step and layer to start TIC is set to $t_0\!=\!4$ and $l_0\!=\!10$, respectively.

\textbf{Baselines and dataset.} For reconstruction and editing, we compare TIC with the following baselines: \textbf{1)} VAE~\cite{rombach2022high}, which directly decodes the latent of the input image without DDIM inversion or sampling, and is commonly considered as the upper bound of reconstruction for LDMs. \textbf{2)} DDIM~\cite{song2020denoising}. \textbf{3)} SDEdit~\cite{meng2021sdedit}, which is an image-to-image method with the strength value set to 0.5. \textbf{4)} Prompt-to-Prompt (P2P)~\cite{hertz2022prompt}, for which we adopt the attention refinement controller to perform non-rigid editing of real images. \textbf{5)} Null-Text Inversion (NTI)~\cite{mokady2022null}, which learns the structural information of the input image into the unconditional embedding of CFG to maintain layout consistency. The total number of fine-tuning iterations for the unconditional embedding is set to 250 or 500 (\emph{i.e.}, 5 or 10 for each sampling step). \textbf{6)} Prompt-Tuning Inversion (PTI)~\cite{dong2023prompt}, which is similar to NTI, but fine-tunes the conditional embedding of CFG instead of the unconditional one. The total number of tuning iterations is also set to 250 or 500 (\emph{i.e.}, 5 or 10 for each sampling step). 7) MasaCtrl~\cite{cao2023masactrl}, which is also tuning-free and can perform non-rigid editing as our method. Unlike our method, it queries contents from the naive DDIM reconstruction process of the source image. Other hyper-parameters of these methods are set to their default values.


For the dataset, we evaluate the reconstruction quality of VAE, DDIM, NTI, PTI and our method on 200 randomly selected images from the MS-COCO 2017 validation set~\cite{lin2014microsoft}. As both NTI and PTI require fine-tuning of the text embedding, we randomly choose one out of five captions for each image from the MS-COCO dataset as the input text prompt.
For editing, we perform consistent editing of real images obtained online.






\subsection{Comparisons on Content-Consistent Image Editing}
\label{sec: 4.2 consistent image editing results}

\textbf{Reconstruction quality.} We first quantitatively evaluate the reconstruction quality of different inversion-based methods on 200 randomly selected images from the MS-COCO validation set. We measure the reconstruction quality by Peak Signal-to-Noise Ratio (PSNR) and Structural Similarity (SSIM), and efficiency by reconstruction time (Time).
As provided in Table~\ref{table: reconstruction PSNR}, the reconstruction quality of our method is significantly superior to DDIM reconstruction, attaining a level of reconstruction that is comparable to VAE, which serves as an upper bound for reconstruction. In addition, compared to the tuning-based methods (\emph{i.e.}, NTI and PTI), our method is tuning-free and much more superior, in terms of both the image reconstruction quality and time.

\textbf{Consistent image editing. }In Fig.~\ref{fig: consistent_all}, we compare the proposed TIC with the baselines on consistent editing of real images. Our method can perform non-rigid edits to the postures (the 1st and 2nd rows), facial expressions (the 3rd row) or certain attributes (the 4th row), while completely preserving the foreground and background contents of the original image, while other methods cannot achieve both. For example, in the 1st row, DDIM can generate results that comply with the target prompt $\mathcal{P}^*$ (\emph{i.e.}, make the person raise his right hand), but the background details are significantly changed. SDEdit and P2P neither make the person raise his hand, nor generate contents that are consistent with the input. This is because they try to keep the original layout or the object shape unchanged by leveraging the layout information encoded in the cross-attention maps (P2P) or the input image (SDEdit). Compared to SDEdit and P2P, NTI and PTI can preserve more content and identity of the original image, as they learn the structural information of the input by fine-tuning the unconditional (NTI) or conditional (PTI) embedding of the classifier-free guidance. However, they still introduce artifacts into the background, and fail to perform non-rigid edits. Finally, MasaCtrl can make the person raise his right hand, but it introduces artifacts in the background (see the areas in the red circles), which disrupts consistency. Since DDIM does not always guarantee perfect reconstruction, querying the contents from the reconstruction process of the source image may lead to unsatisfactory results. 

Our method can also perform consistent editing when there are multiple objects in the input image. For example, for the results in the 4th row of Fig.~\ref{fig: consistent_all}, our method can turn half a glass of milk into a full glass, while keeping the biscuit and the desk surface unchanged. DDIM, P2P and MasaCtrl can also generate a full glass of milk, but they either significantly change the original content, or introduce a large amount of artifacts. SDEdit, NTI and PTI, on the other hand, completely fail to turn the half glass of milk into a full one. More results are included in the supplementary material. 




\subsection{Extensions for General Text-Guided Editing}
\label{sec: 4.3 extented results}

\begin{figure}[t]
    \centering
    \includegraphics[width=\linewidth]{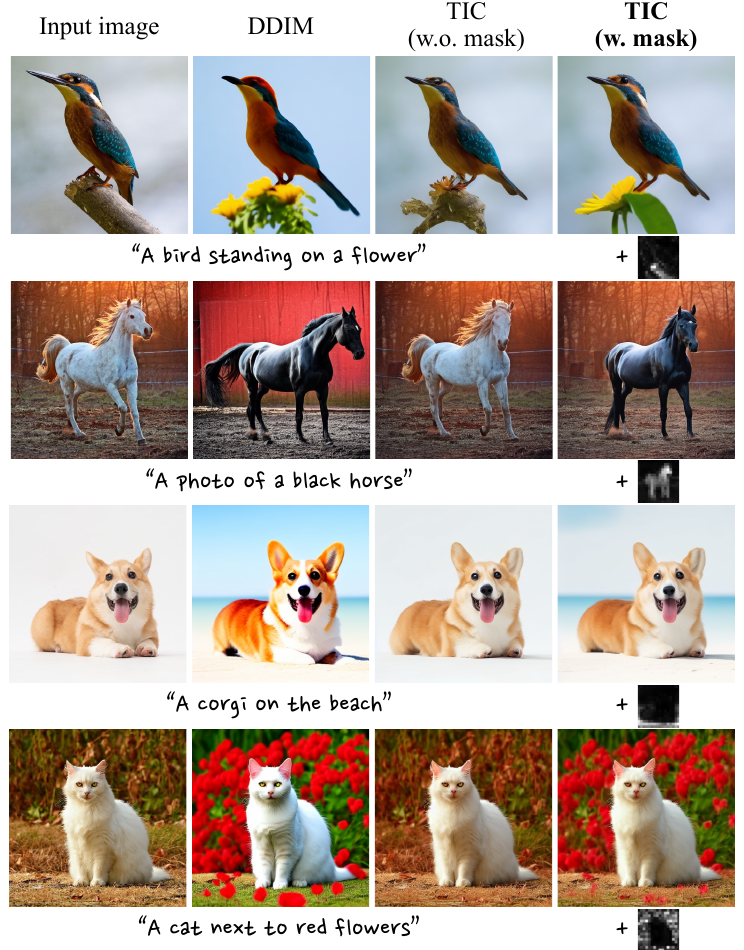}
    \vspace{-5.5mm}
    \caption{Text-guided image editing results of the proposed mask-guided TIC with attention concatenation, which better preserves the contents in areas that do not require editing.}
    \label{fig: mask_aware_IELD_results}
    \vspace{-4mm}
\end{figure}


\textbf{Results of mask-guided TIC with attention concatenation.} We evaluate the performance of the mask-guided TIC with attention concatenation to demonstrate its effectiveness in more general editing settings. We provide editing results in Fig.~\ref{fig: mask_aware_IELD_results} of our method and naive DDIM editing. Under each row, we provide the editing prompt and the cross-attention map for generating the guiding mask, which distinguishes the parts to be edited from the parts to be preserved.

As analyzed before, by only querying contents from inversion of the input, TIC can hardly generate new contents that do not exist in the input (see the 3rd column of Fig.~\ref{fig: mask_aware_IELD_results}). With the proposed mask-guided attention concatenation strategy, TIC
successfully generates new textual contents in $\mathcal{P}^*$ while better preserving the original contents in terms of parts that do not require editing (the 4th column). For example, for images in the 1st and 2nd rows of Fig.~\ref{fig: mask_aware_IELD_results}, the mask-guided TIC successfully edits the target objects (\emph{i.e.}, turning a branch into a flower for the 1st row, and a white horse into a black one for the 2nd row) while preserving the details in other parts. DDIM, on the other hand, can barely retain the details of the input for the unedited parts. Besides, when editing the background of the images in the 3rd and 4th rows, DDIM changes the attributes of the foreground drastically, while our method preserves the appearance of the foreground. From the results, TIC is capable of editing both the foreground and the background thanks to the mask for guidance. The results demonstrate that querying the contents from inversion plays an important role in reconstructing the contents in areas that do not need editing.


\begin{figure}[t]
    \centering
    \includegraphics[width=\linewidth]{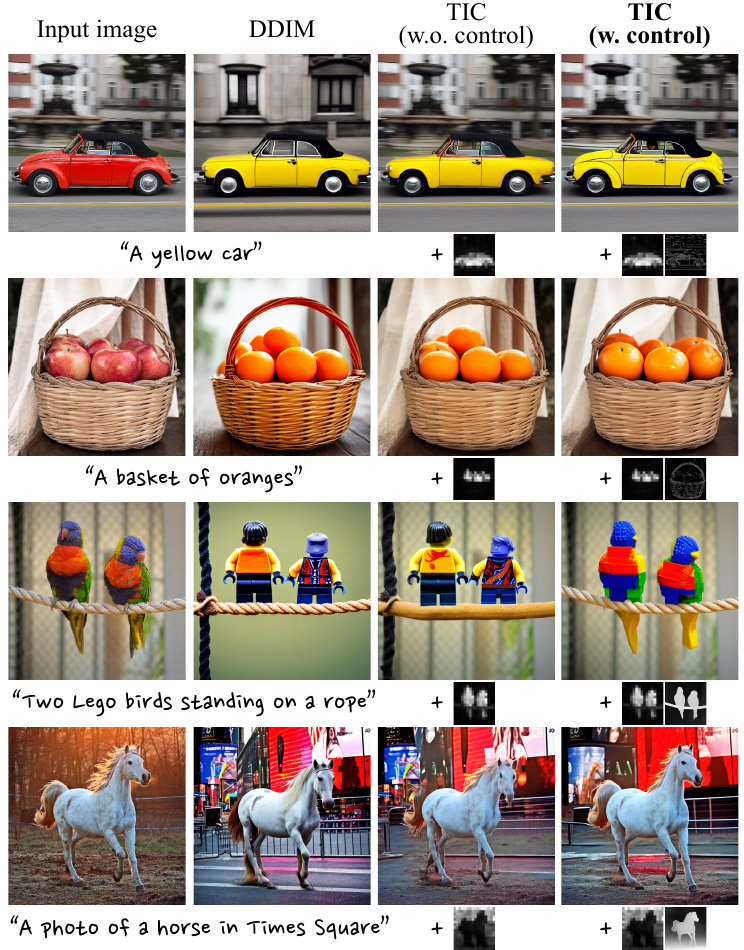}
    \vspace{-5.5mm}
    \caption{Text-guided image editing results of DDIM and the proposed mask-guided TIC (w. or w.o. additional control). By integrating the mask-guided TIC with canny or depth control, our method further enhances fidelity of the editing to both the contents and the layout of the input image.}
    \label{fig: ControlNet_IELD_results}
    \vspace{-4mm}
\end{figure}


\textbf{Results with ControlNet. }By integrating the mask-guided TIC with attention concatenation proposed above into controllable diffusion models (\emph{e.g.}, ControlNet~\cite{zhang2023adding}), our method can further enhance fidelity by preserving the layout and shape of the original image. In our experiment, we adopt the canny or depth map to extract layout information from the input. In Fig.~\ref{fig: ControlNet_IELD_results}, we show qualitative results with ControlNet. Under each row, we also provide the target prompt, the cross-attention map for generating the guiding mask, and the canny map or depth map for ControlNet (only for the results in the 4th column).


From the results, we observe that DDIM significantly altered the details of the original image. For example, for the results in the 1st row, the background of the synthesized image by DDIM (the 2nd column) is largely different from that in the input. With the mask-aware attention concatenation strategy, TIC (the 3rd column) almost perfectly preserves the details of the background, but the shape of the generated yellow car in the foreground is slightly different from that in the input (\emph{i.e.}, the body of the generated car is a little longer). By introducing canny control, TIC (the 4th column) further ensures the consistency of the car's shape and other attributes, only changing its color. The editing results in other rows show similar trends. The results demonstrate that with the combination of this mask-guided attention concatenation strategy and additional control maps, our method effectively combines the layout synthesized by ControlNet with the target prompt and the content in the input image, further enhancing fidelity without losing editability.


\vspace{-2mm}
\section{Conclusion}

In this paper, we propose Tuning-free Inversion-enhanced Control to perform consistent editing of real images. By introducing features from the DDIM inversion process into the sampling process, our method outperforms previous ones in both accurate reconstruction and content-consistent editing of real images. We demonstrate the versatility and applicability of TIC in various experimental settings. Our approach efficiently and effectively tackles the challenges in non-rigid consistent editing, and we believe it will be a valuable tool for more applications in both image and video generation, which we point to as future work.

\section{Acknowledgments}

This research was supported by Zhejiang Provincial Natural Science Foundation of China under Grant No. D24F020011, Beijing Natural Science Foundation L223024, National Natural Science Foundation of China under Grant 62076016 and Grant 92370114. The work was also supported by the National Key Research and Development Program of China (Grant No. 2023YFC3300029) and “One Thousand Plan” projects in Jiangxi Province Jxsg2023102268 and ATR key laboratory grant 220402.

\bibliography{aaai24}

\end{document}